\def\year{2021}\relax
\documentclass[letterpaper]{article} % DO NOT CHANGE THIS
\usepackage{aaai21}  % DO NOT CHANGE THIS
\usepackage{times}  % DO NOT CHANGE THIS
\usepackage{helvet} % DO NOT CHANGE THIS
\usepackage{courier}  % DO NOT CHANGE THIS
\usepackage[hyphens]{url}  % DO NOT CHANGE THIS
\usepackage{graphicx} % DO NOT CHANGE THIS
\usepackage{natbib}  % DO NOT CHANGE THIS AND DO NOT ADD ANY OPTIONS TO IT
\usepackage{caption} % DO NOT CHANGE THIS AND DO NOT ADD ANY OPTIONS TO IT
\graphicspath{ {figures/} }
\usepackage{subcaption}
\usepackage{stix}
\usepackage{pifont}
\usepackage[normalem]{ulem} % underline with \uline

\title{Learning to Control Complex Robots Using High-Dimensional Interfaces: Preliminary Insights}
\author {
	Jongmin M. Lee\textsuperscript{\rm 1}, 
	Temesgen Gebrekristos\textsuperscript{\rm 1},
	Dalia De Santis\textsuperscript{\rm 2},
	Mahdieh Nejati-Javaremi\textsuperscript{\rm 1},
	Deepak Gopinath\textsuperscript{\rm 1},
	Biraj Parikh\textsuperscript{\rm 1},
	Ferdinando A. Mussa-Ivaldi\textsuperscript{\rm 3},
	Brenna D. Argall\textsuperscript{\rm 3}\\
}
\affiliations {
	\textsuperscript{\rm 1,3}Northwestern University, Chicago, Illinois\\
	\textsuperscript{\rm 1,2,3}Shirley Ryan AbilityLab, Chicago, Illinois\\
	\textsuperscript{\rm 2}Istituto Italiano di Tecnologia, Genoa, Italy\\
    \textsuperscript{\rm 1}\{jmlee, tem, m.nejati, deepakgopinath, bparikh\}@u.northwestern.edu, \textsuperscript{\rm 2}dalia.desantis@iit.it, \textsuperscript{\rm 3}\{sandro, brenna.argall\}@northwestern.edu
}

\begin{document}

\maketitle

\begin{abstract}
Human body motions can be captured as a high-dimensional continuous signal using motion sensor technologies. The resulting data can be surprisingly rich in information, even when captured from persons with limited mobility. In this work, we explore the use of limited upper-body motions, captured via motion sensors, as inputs to control a 7 degree-of-freedom assistive robotic arm. It is possible that even dense sensor signals lack the salient information and independence necessary for reliable high-dimensional robot control. As the human learns over time in the context of this limitation, intelligence on the robot can be leveraged to better identify key learning challenges, provide useful feedback, and support individuals until the challenges are managed. In this short paper, we examine two uninjured participants' data from an ongoing study, to extract preliminary results and share insights. We observe opportunities for robot intelligence to step in, including the identification of inconsistencies in time spent across all control dimensions, asymmetries in individual control dimensions, and user progress in learning. Machine reasoning about these situations may facilitate novel interface learning in the future. 

\end{abstract}

\section{Introduction} % (fold)
\label{sec:introduction}

Motion sensor technologies have been used to interface a person's body movements to control machines such as assistive and rehabilitation devices and robots~\cite{RN518, RN455}, drones~\cite{RN878}, and quadcopters~\cite{RN883}. A common strategy for robot control using body motions is to engineer a decoder designed to map the high-dimensional body motion to a lower-dimensional robot control signal space. Whenever the body motion has an intrinsic dimension higher than the device to be controlled, dimensionality reduction techniques, such as principal component analysis (PCA)~\cite{RN884} or autoencoders~\cite{RN885} can be used to implement efficient simultaneous and continuous control of lower-dimensional devices~\cite{pierella2018linear, ranganathan2019age, rizzoglio2021building, thorp2015upper}. However, the design and the operation of such interfaces become challenging when redundancy of the body signals is reduced due to pathological conditions that impact mobility or when controlling complex multi-articulated robotic devices~\cite{chau2017five,RN887}.

These challenges provide a ripe opportunity for robotics autonomy to assist the user~\cite{RN453,RN802,RN897}. For instance, identifying circumstances when a user is control deficient and offering support, may not only benefit both long- and short-term performance, but also help to build trust in assistive and rehabilitation machines~\cite{RN896,RN895}.

In this short paper we present preliminary observations, analyses, and insights on data gathered from two uninjured participants, within an ongoing study, in which a 7 degree-of-freedom (DoF) robot arm is controlled, using a net of sensors on the upper body. Study tasks are designed to familiarize, train, and evaluate robot arm operation via this sensor net, including on Activities of Daily Living (ADL) functional tasks. We describe these experimental methods in the \textsc{Methods} section, share our immediate results in the \textsc{Results} section, and wrap-up our key takeaways and future work in the \textsc{Conclusion and Future Work} section.

% section introduction (end)

\section{Methods} % (fold)
\label{sec:methods}

\begin{figure*}[ht]
    % \captionsetup{belowskip=-6pt}
	\centering
	{\includegraphics[width=1\linewidth]{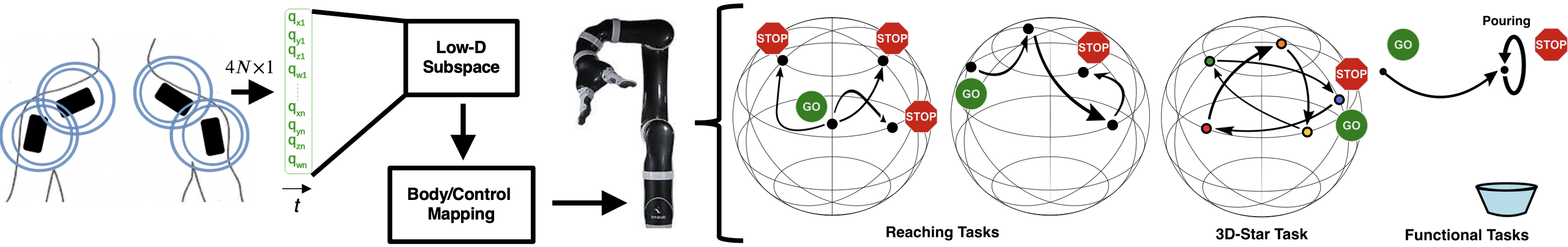}}
	\caption{An overview of the interface-robot pipeline and the study tasks.}
	\label{fig:bomi_system_overview}
\end{figure*}

\noindent{\textbf{Materials.}} The sensor net consists of four inertial measurement unit (IMU) sensors (Yost Labs, Ohio, USA), placed bilaterally on the scapulae and upper arms and anchored to a custom shirt designed to minimize movement artifacts. This is the essence of what is known as the body-machine interface \cite{RN518}. The relative quaternion orientation of the four IMUs in the net (16-dimensional) is mapped to a 6-dimensional subspace using PCA. The PCA map is precomputed using data from an experienced user, performing a predefined set of movements, and this same map is used for all participants. The lower-dimensional subspace consists of 6D velocity commands---3D position ($x,y,z$) and 3D rotation ($\mathit{roll}, \mathit{pitch}, \mathit{yaw}$)---which are used online to control a 7-DoF JACO robotic arm (Kinova Robotics, Quebec, Canada). A GUI is displayed on a tablet to provide a visualization, for the participant, of the robot velocity control commands as well as a score for each trial.

\noindent{\textbf{Protocol.}} There are three phases to the study protocol: (a) familiarization, (b) training, and (c) evaluation (Figure~\ref{fig:bomi_system_overview}). During \textit{familiarization}, participants are encouraged to explore and become familiar with the system on their own, with minimal constraints enforced. Both of the next phases make use of a set of ten fixed targets $\mathcal{G}$. During \textit{training}, two categories of reaching tasks are employed: reaches from a fixed center position out to a target $g_i\in \mathcal{G}$, and sequential reaches between multiple targets $g_j\in\mathcal{G}$. The ordering of targets is random and balanced across days to avoid ordering effects, and it is identical across participants. The \textit{evaluation} phase is split into a reaching and a functional task. In the reaching task, participants reach to five targets that comprise a 3D-star $g_k \in \mathcal{G}$ in fixed succession. The functional tasks are designed to emulate four ADL tasks: (a) take a cup (upside-down) from a dish rack and place it (upright) on the table, (b) pour cereal into a bowl, (c) scoop cereal from a bowl, and (d) throw away a mask in the trash bin.

A trial ends upon successful completion or timeout. For reaching any target $g\in \mathcal{G}$, success is defined within a strict positional (1.00 cm) and rotational (0.02 rad, or 1.14\textdegree) threshold, and the timeout is 1.5 minutes. For the functional tasks, experimenters follow codified guidelines to determine when the tasks complete and the timeout is 3 minutes. Participants are informed of the timeouts and asked to perform tasks to the best of their ability. If there is any risk of harm to the participant or the robot, study personnel intervene and teleoperate the robot to a safe position before proceeding.

% \subsection{Participants} % (fold)
% \label{sub:participants}

\noindent{\textbf{Participants.}} Each participant completes five sessions, executed on consecutive days for approximately two hours each. All sessions are conducted with the approval of the Northwestern University IRB, and all participants provide their informed consent. Two uninjured participants from this on-going study are reported in this paper. P1 is a 31-year-old male, and P2 is a 29-year-old female; both participants are right-handed.

% subsection participants (end)

% section methods (end)

\section{Results} % (fold)
\label{sec:results}

\begin{figure*}[ht]
    \captionsetup{aboveskip=-2pt}
	\centering
	\begin{subfigure}[b]{\textwidth}
		\centering
    	\includegraphics[width=1\linewidth]{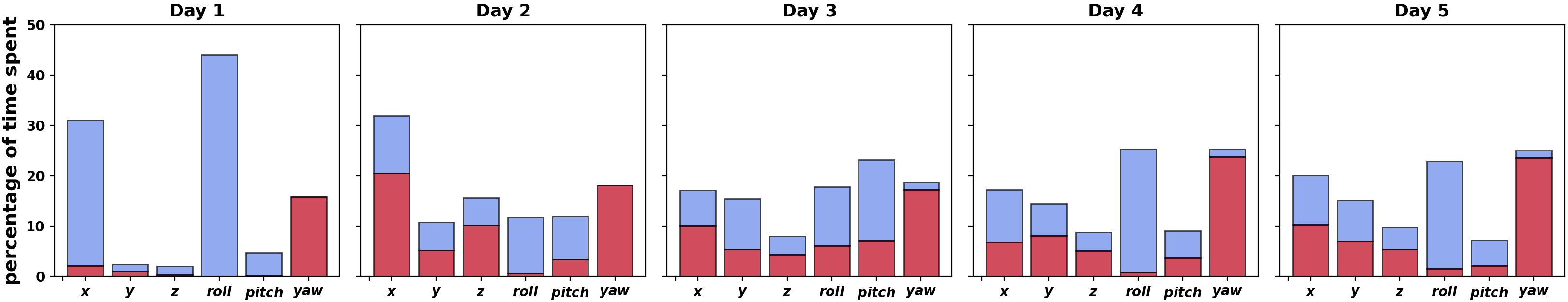}
        \label{fig:p1_proportion_time_spent}
  	\end{subfigure}%
  	\vfill
  	\begin{subfigure}[b]{\textwidth}
		\centering
        \includegraphics[width=1\linewidth]{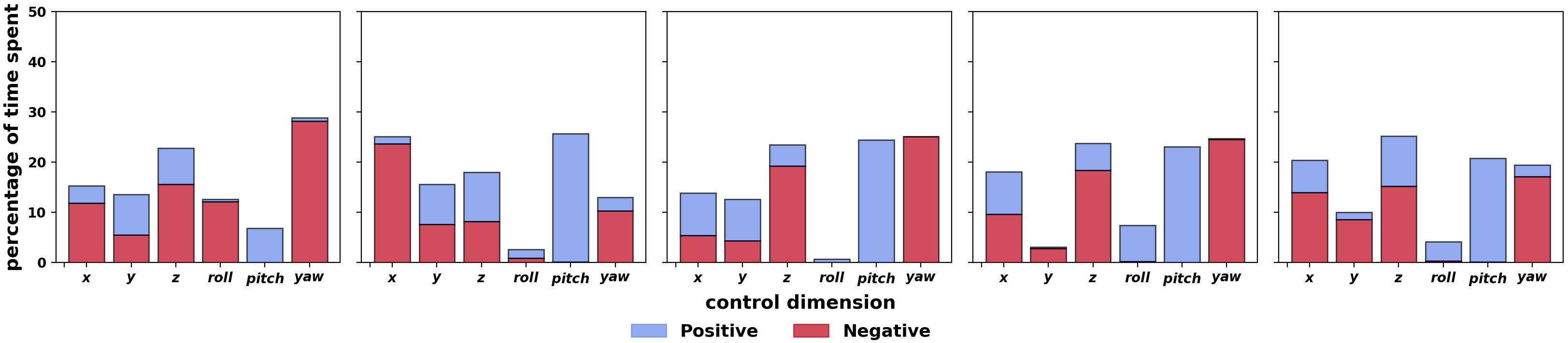}
        \label{fig:p2_proportion_time_spent}
  	\end{subfigure}
	\caption{Five-day evolution of proportion of time spent, in each control dimension, for participants P1 (top) and P2 (bottom) performing the 3D-star task. (Zero commands not included.)}
	\label{fig:proportion_time_spent}
\end{figure*}

\noindent{\textbf{Control Access and Asymmetries.}} We characterize the two participants' control access by tracking the time spent moving the robot in each of the task space control dimensions ($x,y,z$, $\mathit{roll}, \mathit{pitch}$, $\mathit{yaw}$) during the 3D-star task. The percentage of time spent along positive and negative directions of each of the six observed control dimensions is shown in Figure~\ref{fig:proportion_time_spent}. 

P1 initially, on Day 1, spends a majority of time in only two control dimensions (31\% in $x$, and 44\% in $\mathit{roll}$); this is not as apparent for P2. Access to control dimensions, for both participants, is generally asymmetric, with each participant largely accessing either positive direction or negative direction of each control dimension.

\begin{figure*}[ht]
	\centering
  	\begin{subfigure}[b]{0.245\textwidth}
		\centering
    	\includegraphics[width=\linewidth]{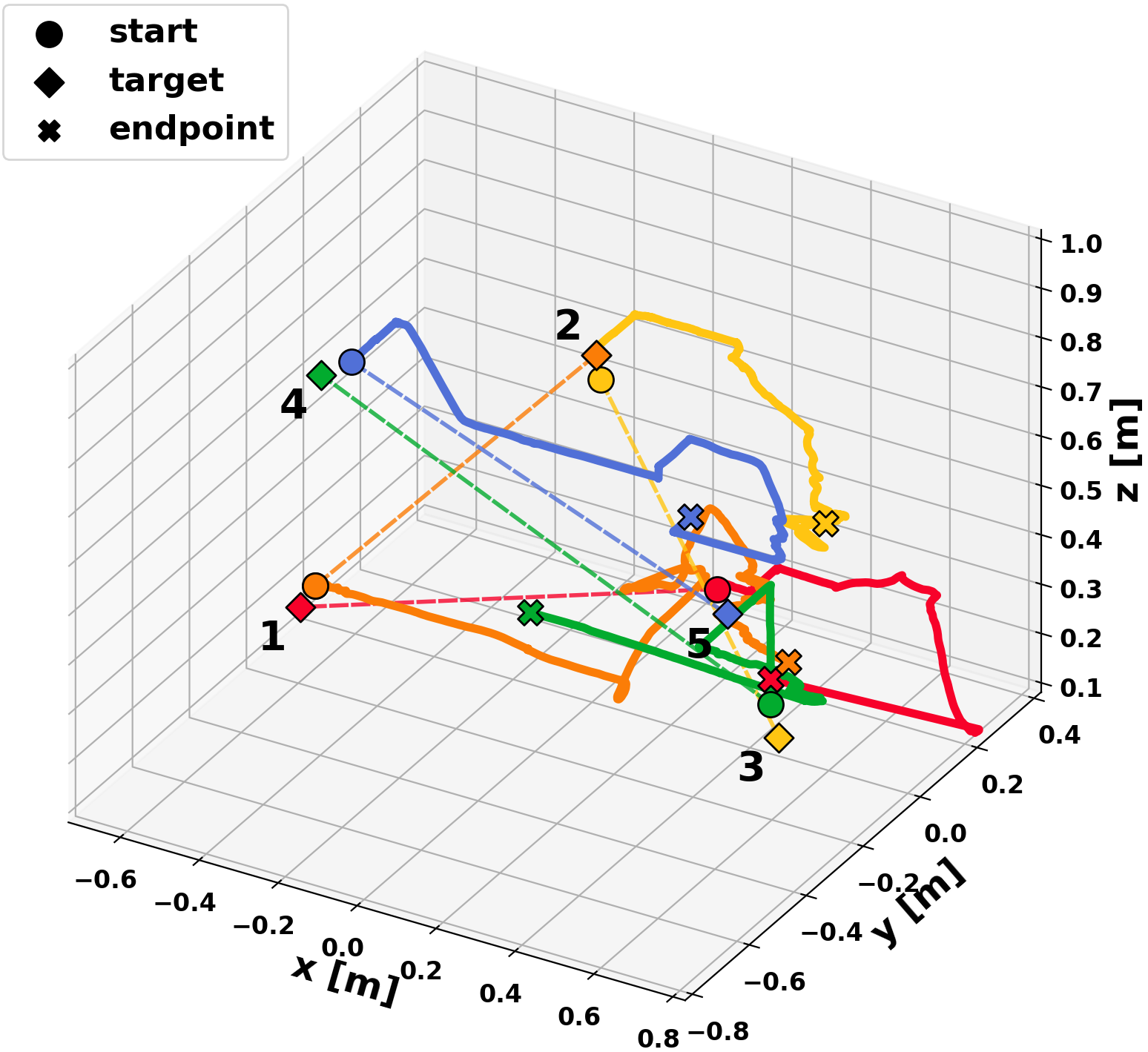}
    	\caption{P1; Day 1} \label{fig:3d_trajectories_a}
  	\end{subfigure}% 
  	\begin{subfigure}[b]{0.245\textwidth}
		\centering
    	\includegraphics[width=\linewidth]{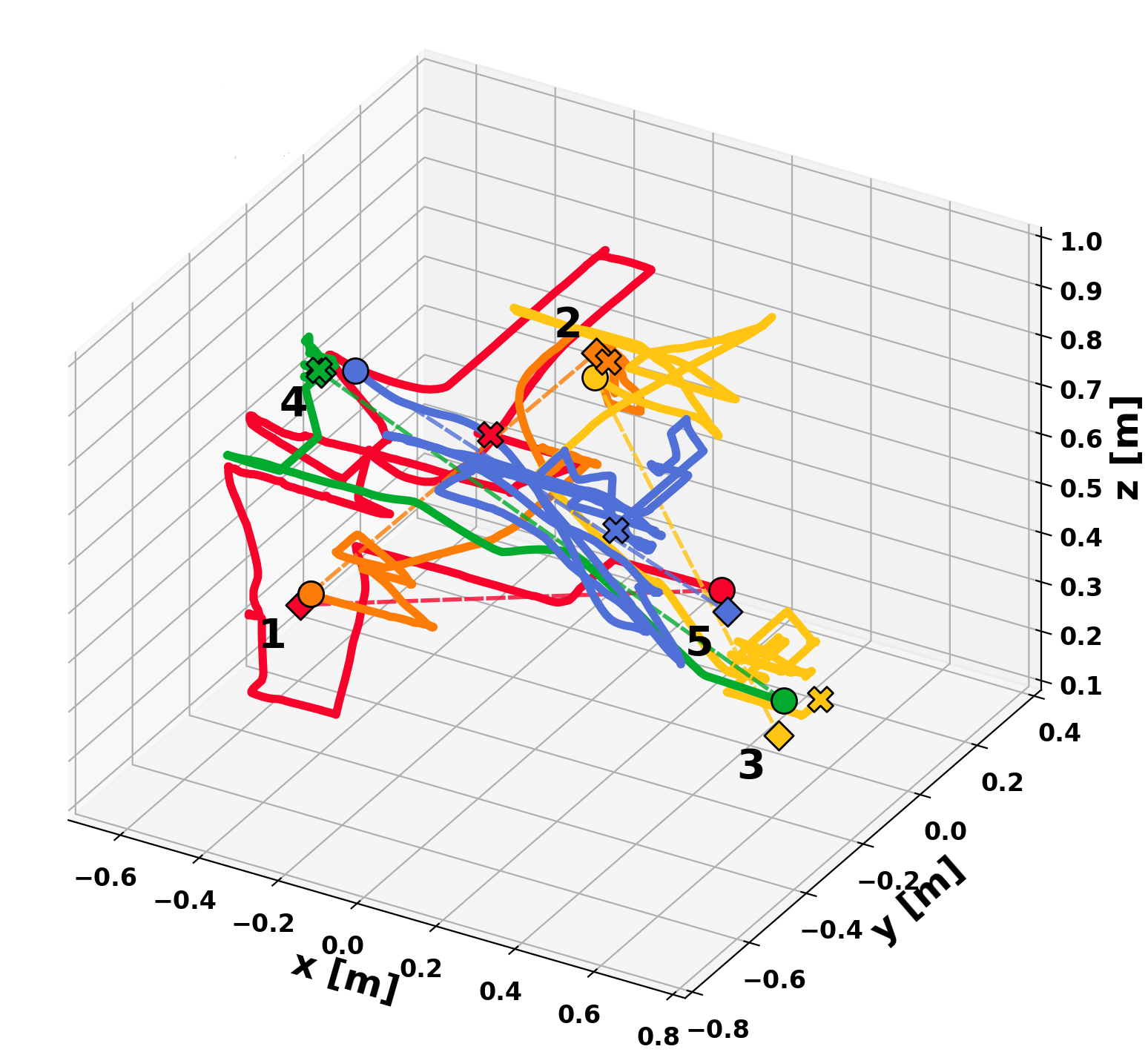}
    	\caption{P1; Day 5} \label{fig:3d_trajectories_b}
  	\end{subfigure}
  	\begin{subfigure}[b]{0.245\textwidth}
		\centering
    	\includegraphics[width=\linewidth]{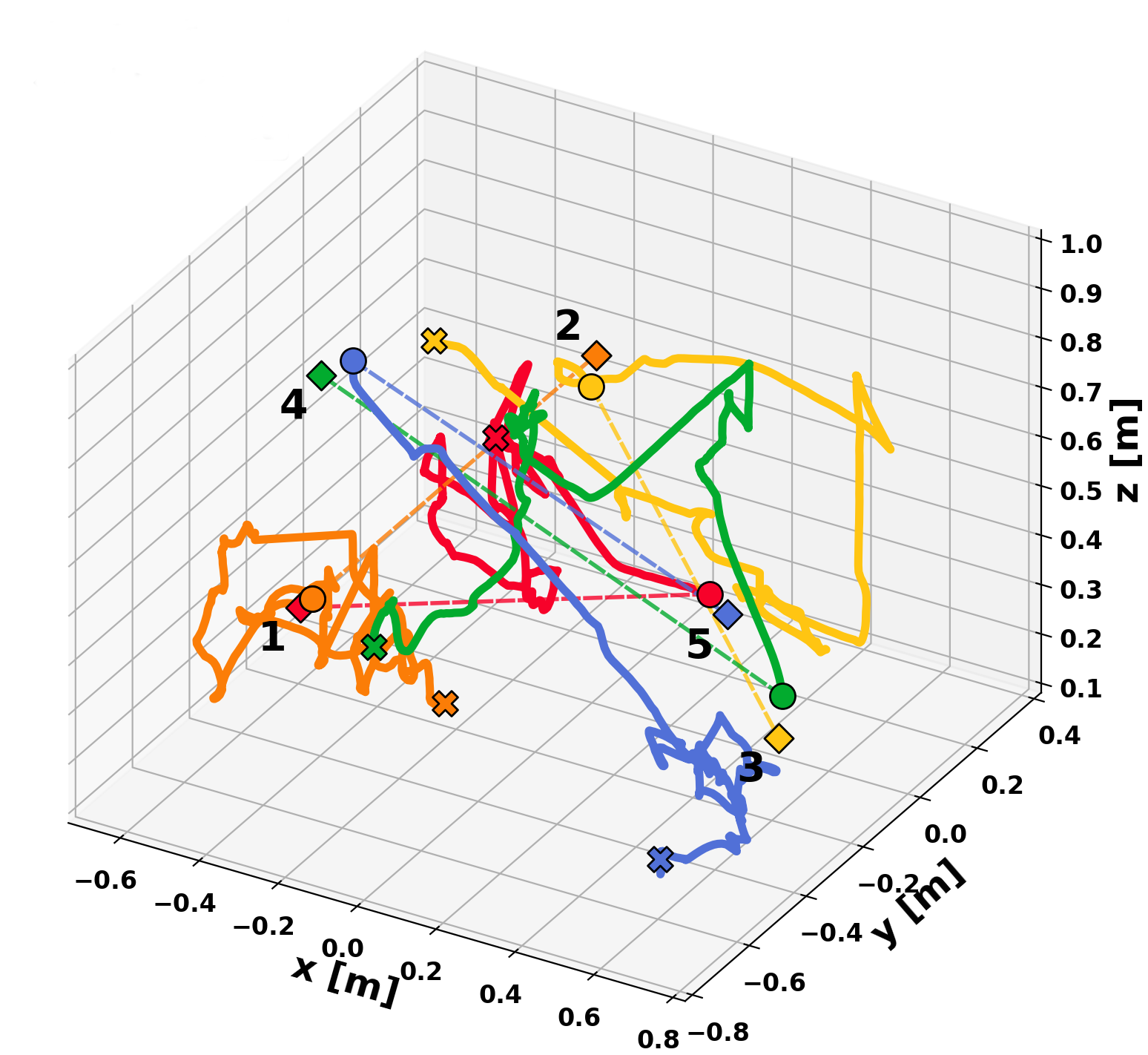}
    	\caption{P2; Day 1} \label{fig:3d_trajectories_c}
  	\end{subfigure}%
	\begin{subfigure}[b]{0.245\textwidth}
		\centering
		\includegraphics[width=\linewidth]{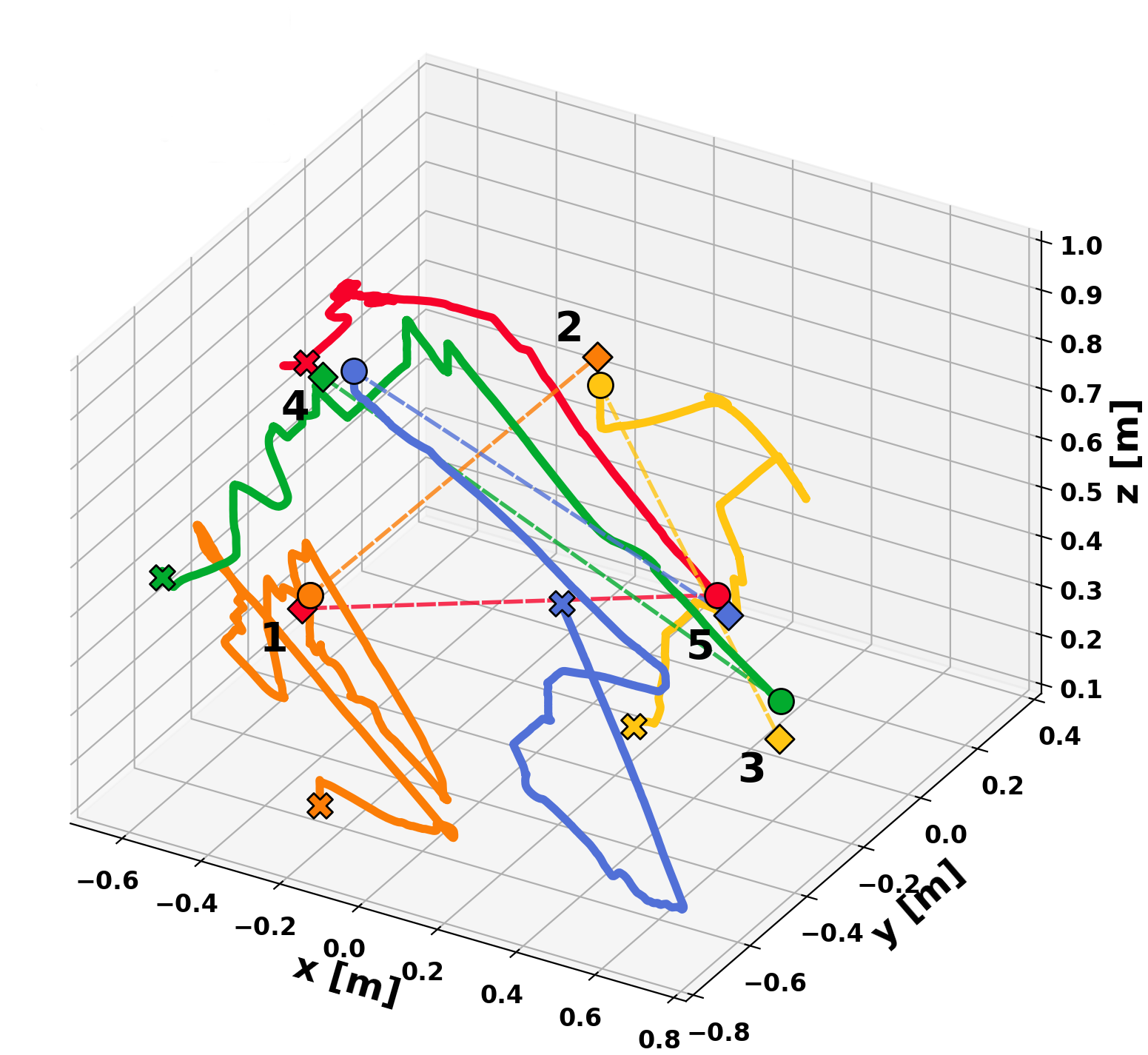}
		\caption{P2; Day 5} \label{fig:3d_trajectories_d}
	\end{subfigure}%
	\caption{Trajectory plots of robot end-effector position during the 3D-star task on Days 1 and 5, for both participants. The task consists of reaching to five different targets ($\mdblkdiamond$) in succession. Start (\ding{108}) and end (\ding{54}) points for each reach, and the straight-line path between them (dotted line), are shown. Each target, start, end, and straight-line path for a single reach are the same color.} \label{fig:3d_trajectories}
\end{figure*}

Over time, the distribution of control access tends to equalize across dimensions for P1 only. The evolution is not necessarily smooth, as swift changes can be observed between consecutive days (Day 1 $\rightarrow$ Day 2). The final distributions themselves differ markedly between the two participants. Most striking is the difference in $y$ and $\mathit{roll}$ access on Days 4 and 5. Recall that the task itself is identical for each participant, and the workspace is obstacle-free. While it is possible that the paths planned by each participant would differ even under perfect control execution, most likely spurious movements are happening as the participants learn the control mapping and interface operation. (This is further supported by the differences in the end-effector trajectories depicted in Figure~\ref{fig:3d_trajectories}.)

From Figure~\ref{fig:proportion_time_spent}, we can also observe that the evolution of access asymmetries follows a distinct pattern for P1 and P2. P1 tends to reduce access asymmetries within a given dimension, as access of positive and negative commands becomes more balanced for all dimensions, as early as Days 2 and 3. However, for P2, only some control dimensions become more balanced over time ($x$, $z$), while others maintain asymmetry (all rotational dimensions) or become more asymmetric ($y$). In addition, the direction of the bias (positive versus negative) is not always consistent. 

\begin{figure*}[ht!]
    \captionsetup{aboveskip=-2pt}
    \centering
    \begin{subfigure}[t]{\textwidth}
		\centering
    	\includegraphics[width=1\linewidth]{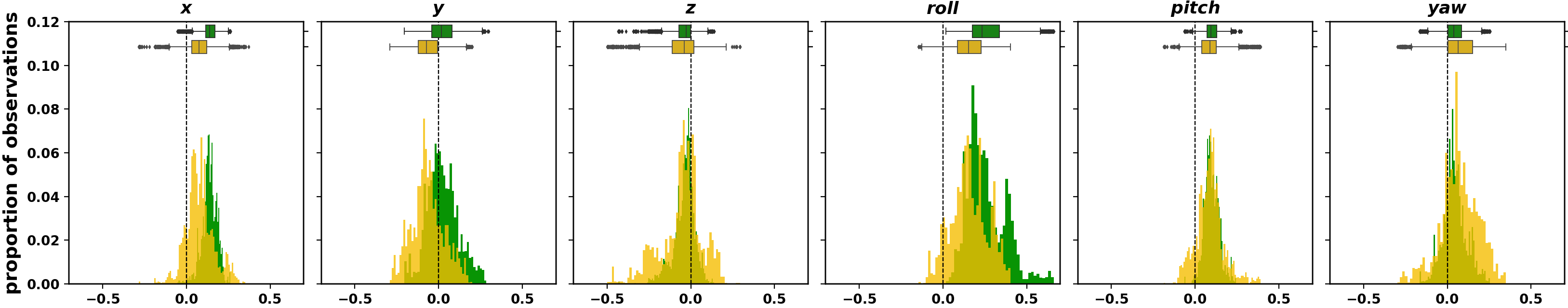}
    \label{fig:applied_commands_a}
  	\end{subfigure}%
  	\vfill
  	\begin{subfigure}[t]{\textwidth}
		\centering
    \includegraphics[width=1\linewidth]{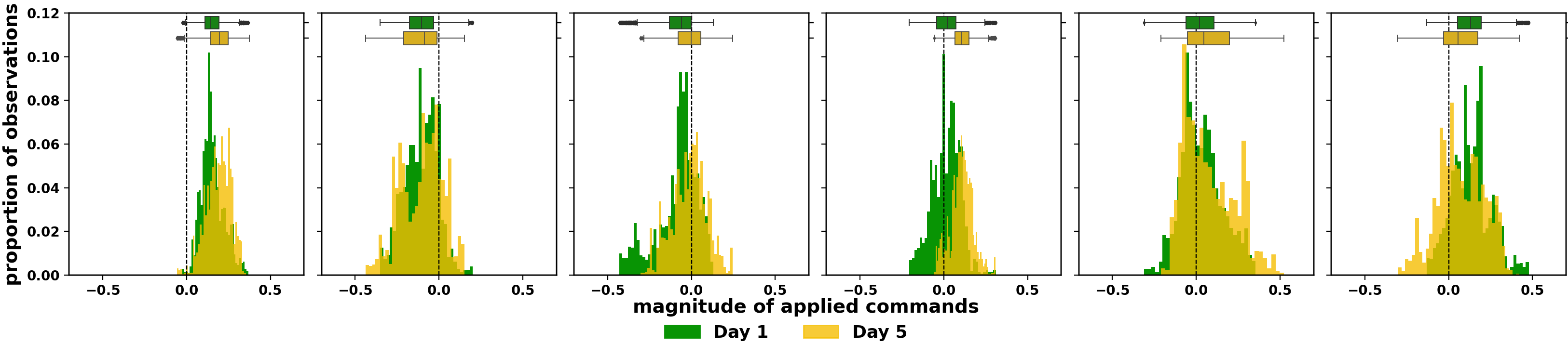}
    \label{fig:applied_commands_b}
  	\end{subfigure}
	\caption{First and last day comparison of histograms of observed robot commands within each control dimension, as each participant (P1 top, P2 bottom) executes the 3D-star task. Standard box plots of robot commands are presented above each respective histogram.} 
	\label{fig:applied_commands}
\end{figure*}

We further examine the distribution of directional access for each control dimension, along with the command magnitude, in Figure~\ref{fig:applied_commands}. Between Days 1 and 5, the histogram supports generally widen, and, with the exception of $\mathit{roll}$, each control dimension, furthermore, exhibits an increase in variance between Days 1 and 5. Each of these trends are visible in the box plots and the changes in the first- and second-order statistics (mean and variance) of the observed commands between days.

\noindent{\textbf{Task Performance.}} To better visualize human learning over the multiple study sessions, we plot the robot end-effector position on Days 1 and 5 in Figure~\ref{fig:3d_trajectories}. In general, for both participants, movements become more successful in reaching (or reaching closer to) the target, more directed (closer to the shortest path), and temporally front-loaded, with the bulk of the distance traveled occurring early. Much of the execution time is spent either in late-execution recovery or in the achievement of final orientation, which requires fine-motor commands (e.g.,~target 5 on Day 5, P1). P2 sometimes reaches near targets early, and then falls into traps of recovery because of P2's difficulty with issuing $+y$ (needed for targets 2 and 4) and $-z$ (needed for target 1) commands. Each participant, furthermore, exhibits an increase in the number of control commands (a decrease in zero commands) between the first and last days, observed in Figure~\ref{fig:3d_trajectories} and from the raw counts of robot commands (not shown). There is also generally a trend of increasing variance across control dimensions between Days 1 and 5; whether this trend is a sign of increased control access and learning or from spurious movements is presently unclear, however.

\noindent\textbf{Opportunities for Robotics Intelligence.} Learning to control complex robots using novel high-DoF interfaces presents many challenges that can be mitigated with the support of robotics intelligence that is designed to be aware and adaptive to the user. Such intelligence might compensate for characteristics of control asymmetries or deficiencies, inconsistencies in time spent across control dimensions, or short- and long-term interface learning, for instance. 

Although not presented in this short paper, subjective feedback gathered via questionnaire indicates that the robot control is unintuitive at times. There are instances when participants feel uncertain about how to move the robot in certain dimensions, despite having become familiar with the dynamics of the robot, and other instances where slight differences in a participant's movements lead to the robot moving in unexpected directions. As a result, we observe participants regularly issuing unintended commands through the interface---either by moving in the undesired direction of an intended control dimension or activating an unintended dimension altogether. The result is time spent attempting corrections and recovery instead of progressing towards task goals. The use of interface-aware autonomy~\cite{RN893} that infers about and prevents these unintended commands in a shared-control framework could not only prevent the subsequent need for corrective action, but can also be used within a training and rehabilitation framework to aid in learning to provide control commands through the interface.  

% section results (end)

 \section{Conclusion and Future Work} % (fold)
 \label{sec:conclusion}

 In this short paper, we presented preliminary results from a study with two uninjured participants in which they controlled a high-DoF robotic arm, using limited upper body movements, to perform a variety of reaching tasks. We presented some key insights on the typical control asymmetries that arise as well as observations on human learning in the context of high-DoF robot control. We also identified intervention opportunities for robotics autonomy. 

 In the future, we will use the data and insights collected to inform the development of an assistive autonomy paradigm. The role of the paradigm will be to facilitate the user's learning of the interface, while adapting to the user's improvement and compensating for any deficits in control. We plan to use the results from this study and the developed autonomy paradigm to conduct a long-term user study, where participants with spinal cord injury evaluate the efficacy of this assistance paradigm.

% section conclusion_future_work (end)

\section{Acknowledgements} % (fold)
\label{sec:acknowledgements}
Research reported in this publication was supported by the National Institute of Biomedical Imaging and Bioengineering (NIBIB) under award number R01-EB024058; National Science Foundation (NSF), award number 2054406; National Institute on Disability, Independent Living and Rehabilitation Research (NIDILRR), award number 90REGE0005-01-00; and European Union's Horizon 2020 Research and Innovation Program under the Marie Sklodowska-Curie, Project REBoT, award number GA-750464. The content is solely the responsibility of the authors and does not necessarily represent the official views of the National Institutes of Health.

% section acknowledgements (end)

%-- References
% \bibliographystyle{aaai21}
\bibliography{aaai-hri-2021}
\end{document}